\pdfoutput=1
\documentclass[10pt,twocolumn,letterpaper]{article}

\usepackage[letterpaper,textwidth=7in,textheight=9in,top=0.75in,%
  hcentering,columnsep=0.33in,footskip=0.45in]{geometry}

\usepackage[T1]{fontenc}
\usepackage{times}
\usepackage{helvet}
\usepackage{courier}
\usepackage{microtype}

\usepackage{amsmath}
\usepackage{amsfonts}
\usepackage{nicefrac}
\usepackage{graphicx}
\usepackage{booktabs}
\usepackage{multirow}
\usepackage{algorithm}
\usepackage{algorithmic}

\usepackage[font=small,labelfont=bf,skip=5pt]{caption}
\usepackage{titlesec}
\usepackage[hyphens]{url}
\usepackage{natbib}
\usepackage{xcolor}

\definecolor{linkblue}{RGB}{0,51,153}
\usepackage[colorlinks=true,linkcolor=linkblue,citecolor=linkblue,%
  urlcolor=linkblue,breaklinks=true,bookmarksnumbered=true]{hyperref}
\hypersetup{
  pdftitle={SharQ: Bridging Activation Sparsity and FP4 Quantization for LLM Inference},
  pdfauthor={Haoqian Meng, Yilun Luo, Yafei Zhao, Wenyuan Liu, Huaqing Zheng, Xindian Ma, Peng Zhang},
  pdfkeywords={FP4 quantization, NVFP4, activation sparsity, N:M sparsity, LLM inference}
}

\titleformat{\section}{\large\bfseries}{\thesection}{0.6em}{}
\titleformat{\subsection}{\normalsize\bfseries}{\thesubsection}{0.6em}{}
\titleformat{\subsubsection}{\normalsize\bfseries}{\thesubsubsection}{0.6em}{}
\titlespacing*{\section}{0pt}{2.0ex plus 0.6ex minus 0.3ex}{1.0ex plus 0.3ex}
\titlespacing*{\subsection}{0pt}{1.6ex plus 0.5ex minus 0.2ex}{0.7ex plus 0.2ex}
\titlespacing*{\subsubsection}{0pt}{1.3ex plus 0.4ex minus 0.2ex}{0.5ex plus 0.2ex}
\makeatletter
\renewcommand{\@maketitle}{%
  \newpage
  \null
  \vskip 0.15in
  \begin{center}%
    {\LARGE\bfseries \@title\par}%
    \vskip 0.22in%
    {\normalsize \lineskip .5em%
      \begin{tabular}[t]{c}\bfseries\@author\end{tabular}\par}%
  \end{center}%
  \par\vskip 0.25in}
\makeatother

\renewenvironment{abstract}%
  {\begin{center}\large\bfseries\abstractname\end{center}\vspace{-0.8ex}%
   \begin{quote}\small}%
  {\end{quote}\vspace{0.3ex}}

\DeclareMathOperator{\GEMM}{GEMM}

\title{SharQ: Bridging Activation Sparsity and FP4 Quantization\\for LLM Inference}

\author{%
Haoqian Meng$^{1,*}$ \quad
Yilun Luo$^{1,*}$ \quad
Yafei Zhao$^{1,*}$ \quad
Wenyuan Liu$^{1}$ \\[2pt]
\textbf{Huaqing Zheng}$^{1}$ \quad
\textbf{Xindian Ma}$^{1}$ \quad
\textbf{Peng Zhang}$^{1,\dagger}$ \\[4pt]
$^{1}$School of Computer Science and Technology, Tianjin University \\[2pt]
{\normalfont\small $^{*}$Equal contribution. \quad $^{\dagger}$Corresponding author: \href{mailto:pzhang@tju.edu.cn}{\texttt{pzhang@tju.edu.cn}}}
}

\date{}

\begin{document}

\maketitle

\begin{abstract}
Low-bit floating-point formats and semi-structured sparsity are increasingly
supported by modern accelerators, yet combining them for LLM activation
compression remains challenging: activations contain input-dependent outliers
that dominate block scales in FP4 quantization, and directly applying N:M
sparsity masks discards moderate values, coupling sparsification loss with
quantization error. We introduce SharQ, a training-free inference method that
bridges activation sparsity and FP4 quantization through an online
sparse--dense decomposition. For each activation tensor, SharQ generates an
input-adaptive N:M mask to extract an outlier-dominated sparse backbone,
quantizes it to FP4, and defines a dense residual relative to the quantized
sparse backbone rather than the unquantized sparse values. A sparse FP4 GEMM
processes the backbone while a dense FP4 GEMM compensates for both
mask-induced activation loss and sparse-path quantization error. The two
paths share a single FP4 weight payload with path-specific scale views, and
a fused preparation kernel absorbs mask generation, residual construction,
and layer normalization into one operator. SharQ requires no calibration
data, retraining, or model-specific tuning. Evaluated on Llama-3.1-8B,
Qwen2.5-7B, Qwen3-30B-A3B, and Qwen3-VL-8B, SharQ recovers 43--63\% of
the NVFP4-to-FP16 accuracy gap across language and vision-language tasks,
and generalizes across NVFP4, HiF4, and MXFP4 formats. On an RTX~5090,
SharQ delivers 2.2--2.4$\times$ latency reduction over FP16 and
1.2--1.4$\times$ throughput improvement over FP8 in language model serving,
and up to 1.58$\times$ speedup on Wan2.2-T2V-A14B video generation when
combined with SageAttention. 
\end{abstract}

\section{Introduction}

Large language model inference is dominated by linear layers. As model size and
serving demand continue to grow, reducing the cost of matrix multiplication has
become a central systems problem. Low-bit floating-point formats and
semi-structured sparsity are two increasingly practical directions, because they
are no longer only algorithmic compression techniques but are also exposed as
hardware execution paths. Block-scaled FP4 formats such as NVFP4, MXFP4, and
HiF4 improve compute density while retaining a floating-point-like dynamic
range, and N:M semi-structured sparsity enables sparse matrix multiplication on
modern Tensor Cores~\citep{nvidia2024nvfp4,
mxspecification, rouhani2023microscalingdataformatsdeep, luo2026hifloat4,
BlackwellArchitectureTechnical}. These trends suggest an appealing target for
LLM inference: combine FP4 quantization with hardware-supported sparse
computation.

However, applying FP4 quantization to activations remains difficult. Compared
with weights, activations are input-dependent and often contain a small number of
large-magnitude outliers~\citep{dettmers2022llmint88bitmatrixmultiplication,
sun2024massiveactivationslargelanguage}. Since block-scaled
FP4 quantization chooses local scales from the largest values in each block,
these outliers can dominate the scale and reduce the effective precision
available to ordinary activation values. Prior work on LLM quantization,
including LLM.int8(), SmoothQuant, AWQ, GPTQ, OmniQuant, QuaRot, and SpinQuant,
has shown from different angles that outliers and activation distributions are
central obstacles for low-bit inference~\citep{dettmers2022llmint88bitmatrixmultiplication,
xiao2024smoothquantaccurateefficientposttraining, lin2023awq,
frantar2023gptqaccurateposttrainingquantization,
shao2024omniquantomnidirectionallycalibratedquantization, ashkboos2024quarot,
liu2024spinquantllmquantizationlearned}. Mixed-precision or outlier-aware paths
can reduce this error, but they often introduce extra storage, irregular data
movement, or additional kernels that are less attractive for an FP4 execution
pipeline.

Activation sparsity offers a tempting alternative. In many LLM layers, the
largest activation values are sparse and carry a disproportionate part of the
output contribution. This suggests that one might route these large values to a
sparse path and exploit N:M sparse Tensor Core instructions. Yet direct
activation sparsification is too lossy in a training-free setting. Unlike weight
sparsity methods such as SparseGPT and Wanda~\citep{sparsegpt, sun2024simple},
where the mask can be optimized or calibrated offline, an activation mask must be
produced online for the current input. Moreover, inference-oriented activation
sparsity methods such as DejaVu and PowerInfer~\citep{liu2023deja,
song2024powerinfer} often rely on coarse-grained prediction or routing, while
fine-grained N:M activation masks must satisfy strict hardware patterns and be
generated cheaply on the critical path. If the sparse path simply keeps the
largest values and drops the rest, many moderate activation values are lost,
leading to substantial accuracy degradation.

The difficulty is therefore not only how to combine sparsity with quantization,
but how to define their roles. In a sparse FP4 path, sparsification changes the
local value distribution before quantization, and the FP4 scale is then selected
from this changed distribution. The error is not the sum of an independent
sparsification error and an independent quantization error: the two are coupled.
A useful design should preserve the hardware benefit of sparse FP4 computation
while retaining the activation information that does not fit into the sparse
representation.

\begin{figure*}[t]
  \centering
  \includegraphics[width=\textwidth]{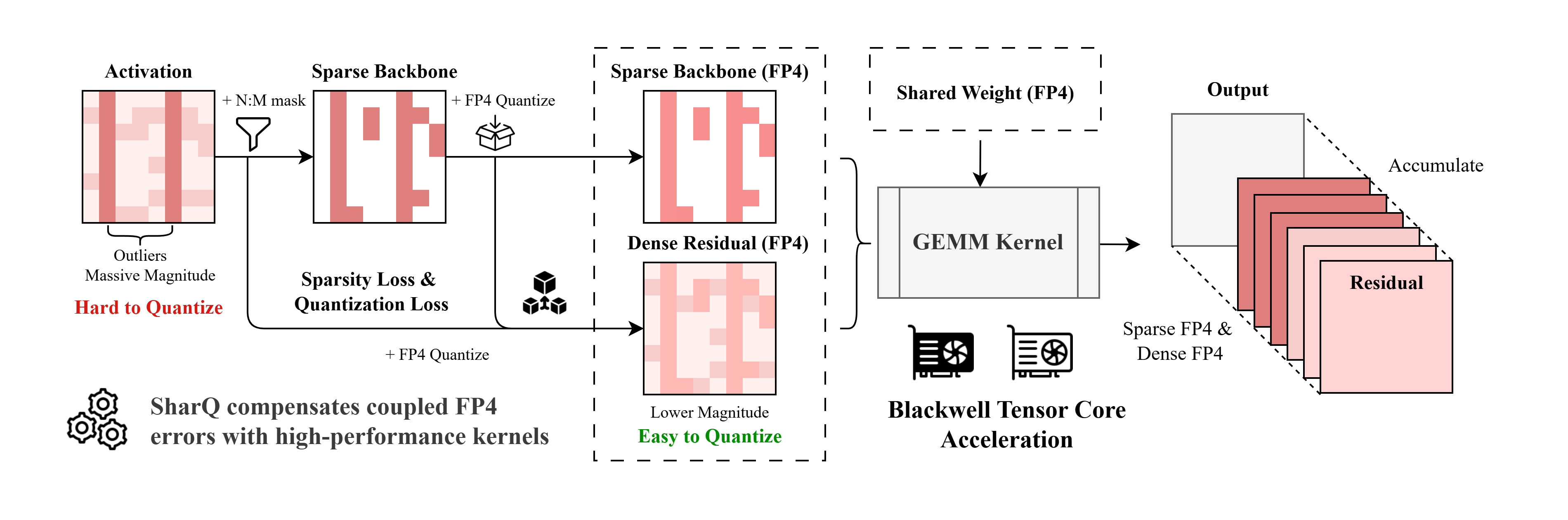}
  \caption{Overview of SharQ. SharQ extracts an outlier-dominated N:M sparse
  backbone, constructs a dense residual relative to the quantized sparse
  backbone, and executes sparse and dense FP4 paths with shared FP4 weights to
  jointly compensate sparsification and quantization errors through
  high-performance kernels.}
  \label{fig:sharq-overview}
\end{figure*}

We introduce SharQ, a training-free inference method that bridges activation
sparsity and FP4 quantization through an online sparse--dense decomposition.
SharQ does not treat activation decomposition itself as the novelty. Instead, it
makes the decomposition input-adaptive, hardware-valid, and quantization-aware.
For each activation, SharQ generates an online N:M mask that extracts an
outlier-dominated sparse backbone. This backbone is quantized and executed by a
sparse FP4 GEMM. SharQ then defines a dense residual relative to the quantized
sparse backbone, rather than relative to the unquantized sparse values. As a
result, the residual contains both the values outside the sparse mask and the
quantization error introduced by the sparse FP4 path. A dense FP4 GEMM then
accumulates this residual contribution into the sparse output.
Figure~\ref{fig:sharq-overview} provides an overview of this pipeline.

This design has two practical consequences. First, the sparse mask is determined
from the current activation, so SharQ does not require calibration data,
finetuning, or a learned offline mask. This helps explain why the same mechanism
works across dense decoder-only models, mixture-of-experts models, video
generation models such as Wan2.2 text-to-video~\citep{wan2025wan}, and
different FP4 formats. Second, SharQ is designed around real kernel constraints:
mask selection, sparse input compression, and residual preparation are fused;
the two paths share the FP4 weight payload with path-specific scale views; and
the dense residual path accumulates directly into the sparse output. Thus, SharQ
is not only an activation approximation strategy, but also a hardware-oriented
inference primitive.

Our contributions are as follows:
\begin{itemize}
  \item We propose SharQ, a training-free FP4 inference method that performs an
  online activation decomposition into an N:M sparse backbone and a dense FP4
  residual.
  \item We define the residual with respect to the quantized sparse backbone,
  allowing the dense path to jointly compensate for mask-induced activation loss
  and sparse-path FP4 quantization error.
  \item We design an efficient implementation with fused activation preparation,
  shared FP4 weights with path-specific scales, and accumulation-based residual
  compensation, and evaluate it across multiple model architectures, FP4
  formats, and inference settings.
\end{itemize}

\section{Related Work}

Prior post-training quantization methods largely address activation outliers by reshaping activation distributions, assigning a subset of channels to higher precision, or directly compensating quantization error. Representative examples include SmoothQuant~\citep{xiao2024smoothquantaccurateefficientposttraining}, AWQ~\citep{lin2023awq}, GPTQ~\citep{frantar2023gptqaccurateposttrainingquantization}, OmniQuant~\citep{shao2024omniquantomnidirectionallycalibratedquantization}, and ARCQuant~\citep{meng2026arcquant}. However, these pipelines still mainly treat activations as dense tensors to be compressed, even in settings where activation outliers are sparse and hardware exposes structured sparse execution.

Weight sparsity methods such as SparseGPT~\citep{sparsegpt} and Wanda~\citep{sun2024simple} learn or calibrate fixed masks offline, while activation sparsity methods such as DejaVu~\citep{liu2023deja} and PowerInfer~\citep{song2024powerinfer} typically rely on coarse-grained or unstructured routing. In contrast, SharQ targets online semi-structured activation sparsity under hardware-valid constraints and couples it with FP4 quantization through a sparse--dense decomposition, so that moderate values are compensated rather than simply dropped.

Recent hardware and algorithmic advances further motivate this setting. Blackwell Tensor Cores natively support NVFP4, MXFP4, and semi-structured sparse execution~\citep{BlackwellArchitectureTechnical,nvidia2024nvfp4}, while recent FP4-oriented methods such as ARCQuant, MR-GPTQ, Four Over Six, and MicroMix~\citep{meng2026arcquant,egiazarian2025bridging,cook2025fouroversix,liu2025micromix} adapt quantization algorithms to these formats. SharQ is complementary to this line of work: instead of redesigning the FP4 codebook or introducing a separate high-precision branch, it reorganizes activations into a sparse FP4 backbone and a dense FP4 residual that remain aligned with hardware-native execution.

\section{Methodology}

\subsection{Preliminary}

Consider a linear layer with activation matrix $\mathbf{X} \in \mathbb{R}^{T \times K}$, weight matrix $\mathbf{W} \in \mathbb{R}^{K \times D}$, and full-precision output
\begin{equation}
  \mathbf{Y} = \mathbf{X}\mathbf{W}.
  \label{eq:linear-layer}
\end{equation}
This work studies how to approximate this computation under hardware-imposed low-bit and sparsity constraints. In group-wise low-bit quantization, a local scale is chosen from the largest magnitude in each group,
\begin{equation}
  s_{\mathcal{G}} =
  \frac{\max_{i \in \mathcal{G}} |x_i|}{\alpha_{\mathcal{B}}},
  \quad
  q_i =
  \Pi_{\mathcal{B}}\left(\frac{x_i}{s_{\mathcal{G}}}\right),
  \quad
  \widehat{x}_i = s_{\mathcal{G}} q_i ,
  \label{eq:group-quant}
\end{equation}
which makes FP4 formats such as NVFP4 sensitive to outliers. Recent hardware also supports N:M semi-structured sparsity, where a dynamic activation mask satisfies
\begin{equation}
  \mathbf{X}_{\mathrm{sp}} =
  \mathbf{M}(\mathbf{X}) \odot \mathbf{X},
  \quad
  \mathbf{M}(\mathbf{X}) \in \mathcal{S}_{N\!:\!M},
  \label{eq:dynamic-mask}
\end{equation}
allowing sparse low-bit execution but coupling sparsification and quantization errors.

\subsection{Motivation}

SharQ is motivated by two observations. First, a few activation outliers can dominate FP4 group scales and sharply reduce the effective precision available to ordinary values. Second, directly applying a dynamic N:M mask is too lossy in a training-free setting: although outliers are sparse and output-critical, many moderate values outside the mask still carry important aggregate information. Once activations are sparsified before quantization, the mask also changes the local value distribution, so sparsification and quantization errors become coupled rather than independent. SharQ therefore reorganizes activations into an outlier-dominated sparse backbone and a dense residual, allowing the dense path to compensate both masked-out values and sparse-path quantization error while staying within hardware-friendly FP4 execution.

\subsection{Decomposition strategy}

At inference time, SharQ performs this decomposition online and reframes
sparsification as routing rather than dropping: outliers are represented by a
sparse FP4 backbone, while the remaining activation information is routed to a
dense FP4 residual. Given an activation matrix $\mathbf{X}$,
SharQ approximates the original linear layer with two complementary paths. The
sparse path captures the outlier-dominated structure under an N:M constraint,
and the dense path preserves the information that cannot be faithfully expressed
by the sparse FP4 representation.

\textbf{Sparse backbone.}
SharQ first generates an online mask from the current activation values. Within
each local block, the mask selects the top-$k$ entries by magnitude, or
equivalently the entries required by the target N:M sparse pattern. This gives an
input-adaptive sparse backbone
\begin{equation}
  \mathbf{X}_{\mathrm{sp}}
  =
  \mathbf{M}(\mathbf{X}) \odot \mathbf{X},
  \quad
  \mathbf{M}(\mathbf{X}) \in \mathcal{S}_{N\!:\!M}.
  \label{eq:sharq-sparse-backbone}
\end{equation}
Here $\mathcal{S}_{N\!:\!M}$ denotes the set of hardware-valid semi-structured
masks. In concrete FP4 formats, SharQ can instantiate this selection with the
corresponding sparse structure, such as 4:8 sparsity in pairs for NVFP4 and the
standard 2:4 pattern for HiF4. Since the mask is computed from the current
activation, the sparse backbone follows input-dependent outlier patterns without
learning a fixed mask offline.

\textbf{Dense residual.}
The sparse backbone is still quantized before entering the sparse path. Let
$\mathcal{Q}_{\mathrm{sp}}$ and $\mathcal{D}_{\mathrm{sp}}$ denote the
quantization and dequantization operators used by the sparse path. The activation
actually represented by that path is
\begin{equation}
  \widetilde{\mathbf{X}}_{\mathrm{sp}}
  =
  \mathcal{D}_{\mathrm{sp}}
  \left(
    \mathcal{Q}_{\mathrm{sp}}(\mathbf{X}_{\mathrm{sp}})
  \right).
  \label{eq:sharq-quantized-backbone}
\end{equation}
SharQ defines the dense residual relative to this quantized sparse backbone:
\begin{equation}
  \mathbf{R}
  =
  \mathbf{X} - \widetilde{\mathbf{X}}_{\mathrm{sp}}
  =
  (\mathbf{X} - \mathbf{X}_{\mathrm{sp}})
  +
  (\mathbf{X}_{\mathrm{sp}} - \widetilde{\mathbf{X}}_{\mathrm{sp}}).
  \label{eq:sharq-residual}
\end{equation}
Thus, $\mathbf{R}$ is not merely the values outside the sparse mask. It also
contains the representation error introduced when the retained outliers are
quantized into FP4. Defining the residual after sparse quantization is crucial:
it lets the dense path compensate both the values excluded by the mask and the
quantization error introduced on the retained sparse backbone. This definition
organizes the sparse approximation loss and the sparse-path quantization loss
into a single residual signal.

\textbf{Residual path compensation.}
SharQ reconstructs the layer output by combining a sparse GEMM on the backbone
and a dense GEMM on the residual:
\begin{equation}
\begin{aligned}
  \widehat{\mathbf{Y}}_{\text{SharQ}}
  ={}& \GEMM_{\mathrm{sp}}\left(
    \mathcal{Q}_{\mathrm{sp}}(\mathbf{X}_{\mathrm{sp}}),
    \mathcal{Q}_{\mathrm{w},\mathrm{sp}}(\mathbf{W})
  \right) \\
  &+ \GEMM_{\mathrm{dn}}\left(
    \mathcal{Q}_{\mathrm{dn}}(\mathbf{R}),
    \mathcal{Q}_{\mathrm{w},\mathrm{dn}}(\mathbf{W})
  \right).
\end{aligned}
\label{eq:sharq-output}
\end{equation}
The first term exploits semi-structured sparse FP4 computation for the
outlier-dominated backbone, while the second term uses dense FP4 computation to
compensate for the information and quantization error left by the sparse path.
Because the residual is defined after sparse quantization, the dense path
compensates both mask-induced approximation loss and sparse-path quantization
loss. SharQ therefore avoids a hardware-unfriendly high-precision outlier path
while retaining the execution advantage of N:M sparse FP4 computation.

\subsection{Kernel design}

\begin{figure*}[t]
  \centering
  \includegraphics[width=\textwidth]{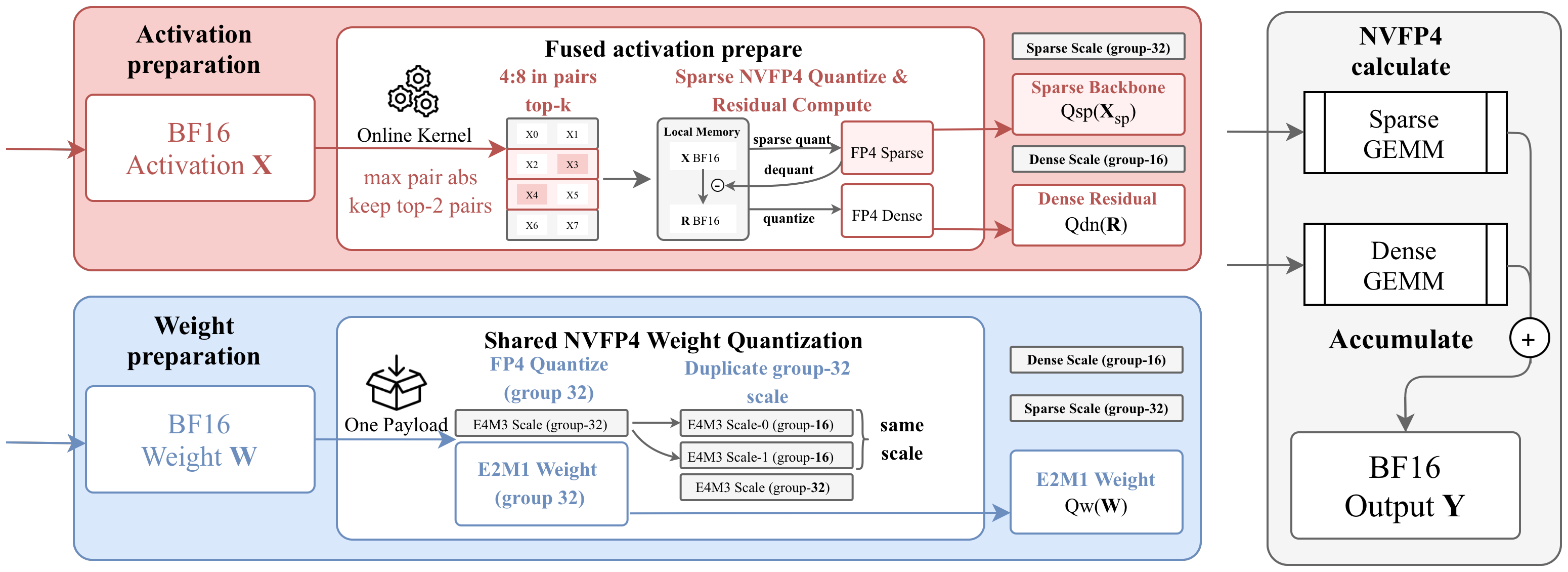}
  \caption{Kernel-level NVFP4 dataflow of SharQ. The fused activation
  preparation kernel performs RMSNorm, pair-topk 4:8 in-pairs mask
  generation, sparse compression, and dense residual construction. Weight
  preparation quantizes a shared NVFP4 weight payload once and exposes
  path-specific scale layouts. The sparse backbone and dense residual are then
  executed by sparse and dense NVFP4 GEMMs, with the residual path accumulating into the output.}
  \label{fig:sharq-kernel-design}
\end{figure*}

The kernel design of SharQ serves one purpose: to preserve the efficiency
promised by activation decomposition under real inference constraints. A
decomposition that improves approximation quality but introduces expensive online
preprocessing, duplicated weight storage, or extra memory traffic would have
limited practical value. SharQ therefore maps the sparse backbone and dense
residual to hardware-compatible low-bit execution paths while minimizing the
coordination overhead between them.
Figure~\ref{fig:sharq-kernel-design} summarizes this kernel-level dataflow.

\textbf{Fused sparse-residual preparation.}
The first efficiency challenge is that dynamic mask generation lies on the
inference critical path. If implemented as a standalone step, its cost can
easily offset the gain from sparse computation. SharQ addresses this by fusing
mask selection with the preparation of both execution paths. In the current
Blackwell NVFP4 implementation, the preparation kernel selects an
outlier-dominated sparse backbone under a local 4:8 in-pairs constraint,
quantizes and compresses it into the input format required by the sparse GEMM,
and simultaneously constructs a dense residual relative to the quantized sparse
backbone. The same preparation pipeline also produces the FP4 residual input for
the dense compensation path. As a result, online mask generation is not exposed
as a separate expensive operator, but absorbed into the activation preparation
stage already required by the two-path computation. For layers with RMSNorm,
SharQ further integrates normalization-aware preprocessing into this
preparation pipeline, reducing additional activation transformation overhead.

\textbf{Shared FP4 weights with path-specific scales.}
A naive two-path implementation would maintain one low-bit weight tensor for the
sparse path and another for the dense path, increasing both storage cost and
weight preparation overhead. SharQ instead keeps a single shared FP4 weight
payload and lets both paths reuse the same low-bit weight body. The difficulty is
that the two paths do not follow the same scale granularity: the dense NVFP4
path uses the standard dense scale organization, whereas the N:M sparse path
follows the scale organization required by the sparse kernel. SharQ resolves this
mismatch without duplicating the FP4 payload itself. It retains path-specific
scale views on top of the shared weight representation, so that both the sparse
path and the dense residual path can use scale layouts compatible with their own
kernels while avoiding the storage and memory overhead of maintaining two
separate quantized weight tensors.

\textbf{Residual path compensation by accumulation.}
SharQ also avoids a naive two-branch output implementation. The sparse backbone
is first processed by a sparse GEMM to produce the main output contribution. The
dense residual path is then applied as a dense NVFP4 GEMM that accumulates
directly into this output, rather than materializing a second full branch result
and merging it afterward. This implementation matches the algorithmic role of
the residual path: it is not an equally weighted parallel branch, but a
compensation path for the information and quantization error left by the sparse
path. More importantly, accumulation-based compensation removes an extra output
writeback and readback, allowing the two-path decomposition to preserve a
dataflow close to that of a single linear kernel. Combined with fused activation
preparation and shared-weight design, this makes SharQ an efficiency-oriented
inference primitive rather than a decomposition that exists only at the
algorithmic level.

\section{Experiments}

\subsection{Experimental setup}

We evaluate SharQ on three representative LLMs spanning different architectures
and scales: Llama-3.1-8B~\citep{grattafiori2024llama3herdmodels} and Qwen2.5-7B~\citep{qwen2025qwen25technicalreport} as standard dense decoder-only models,
and Qwen3-30B-A3B~\citep{qwen3} as a Mixture-of-Experts model with 30B total parameters and
approximately 3B active parameters per token. This selection covers both dense
and sparse-activated architectures.

The primary baselines include RTN-based NVFP4 quantization applied to both
activations and weights, SmoothQuant and ARCQuant under the same NVFP4
setting, and a mixed-format W4A8 baseline with MXFP4 weights and MXFP8
activations. FP16 serves as the full-precision reference. SharQ is applied on
top of RTN-based NVFP4 using the 4:8 in-pairs sparsity pattern supported by
Blackwell Tensor Cores, without any training or calibration data.

We report zero-shot and few-shot accuracy on five standard benchmarks:
ARC-Challenge~\citep{allenai:arc}, HellaSwag~\citep{zellers2019hellaswag}, Lambada~\citep{lambada}, PIQA~\citep{piqa}, and WinoGrande~\citep{sakaguchi2019winograndeadversarialwinogradschema}, along with
their unweighted average. We additionally report perplexity on WikiText2~\citep{wikitext} (lower
is better) and few-shot accuracy on MMLU~\citep{mmlu} as complementary metrics that are
particularly sensitive to quantization degradation. All evaluations are conducted using the Language Model Evaluation Harness~\citep{lm-eval}.

\subsection{Main results}
Table~\ref{tab:main-results} summarizes the main results. Across all three
models, SharQ consistently recovers a substantial portion of the accuracy lost
by NVFP4+RTN quantization. On Llama-3.1-8B, SharQ raises the average zero-shot
accuracy from 70.32\% to 71.55\% and improves MMLU from 61.93\% to 63.76\%. A
similar pattern holds for Qwen2.5-7B (69.43\% $\to$ 70.38\%) and Qwen3-30B-A3B
(68.40\% $\to$ 69.17\%), indicating that the same sparse--dense decomposition
remains effective across both dense and MoE architectures.

\begin{table*}[t]
\caption{Zero-shot and few-shot evaluation of SharQ across three LLMs.
Accuracy (\%) is reported for classification benchmarks; perplexity
($\downarrow$) is reported for WikiText2. The W4A8 baseline uses MXFP4
weights and MXFP8 activations, while RTN, SmoothQuant, and ARCQuant are
reported under the NVFP4 setting.}
\label{tab:main-results}
\centering
\begin{tabular}{@{}l|l|cccccc|c|c@{}}
\toprule
\multicolumn{2}{c|}{} & \multicolumn{6}{c|}{Zero-shot Accuracy ($\uparrow$)} & \multicolumn{1}{c|}{PPL ($\downarrow$)} & \multicolumn{1}{c}{MMLU ($\uparrow$)} \\
Model & Method & ARC-C & Hella & Lamba & PIQA & Wino & Avg. & WikiText2 & 5-shot \\
\midrule
\multirow{6}{*}{Llama-3.1-8B}
 & FP16  & 53.50 & 78.96 & 75.33 & 81.23 & 73.48 & 72.50 & 6.25 & 65.24 \\
 & W4A8 & 50.77 & 78.08 & 73.82 & 79.92 & 71.43 & 70.80 & 6.98 & 62.14 \\[1pt]
 \cline{2-10}
 & RTN & 49.91 & 77.41 & 74.02 & 79.60 & 70.64 & 70.32 & 6.94 & 61.93 \\
 & SmoothQuant & 50.34 & 77.51 & 73.32 & 79.22 & 71.59 & 70.40 & 6.92 & 61.84 \\
 & ARCQuant & 52.73 & 77.17 & 73.76 & 80.41 & 71.11 & 71.04 & 6.88 & 62.41 \\
 & SharQ & 51.96 & 78.40 & 74.67 & 80.20 & 72.53 & \textbf{71.55} & \textbf{6.73} & \textbf{63.76} \\
\midrule
\multirow{6}{*}{Qwen2.5-7B}
 & FP16  & 51.01 & 78.94 & 71.92 & 79.92 & 72.93 & 70.94 & 6.85 & 74.16 \\
 & W4A8 & 50.51 & 77.32 & 68.19 & 79.60 & 70.96 & 69.32 & 7.37 & 72.11 \\[1pt]
 \cline{2-10}
 & RTN & 51.19 & 77.55 & 70.37 & 78.73 & 69.30 & 69.43 & 7.29 & 72.06 \\
 & SmoothQuant & 51.62 & 77.56 & 70.70 & 79.05 & 69.85 & 69.76 & 7.23 & 72.27 \\
 & ARCQuant & 50.51 & 77.81 & 71.63 & 78.78 & 70.56 & 69.86 & 7.28 & 72.12 \\
 & SharQ & 52.73 & 78.08 & 70.00 & 79.43 & 71.67 & \textbf{70.38} & \textbf{7.15} & \textbf{72.83} \\
\midrule
\multirow{6}{*}{\shortstack[l]{Qwen3-30B\\-A3B}}
 & FP16  & 56.31 & 77.72 & 64.86 & 80.58 & 70.88 & 70.07 & 8.70 & 79.64 \\
 & W4A8 & 53.58 & 76.44 & 61.36 & 79.27 & 69.06 & 67.94 & 9.21 & 78.04 \\[1pt]
 \cline{2-10}
 & RTN & 54.52 & 76.76 & 62.91 & 78.35 & 69.46 & 68.40 & 9.12 & 77.72 \\
 & SmoothQuant & 54.27 & 75.73 & 63.61 & 78.02 & 67.96 & 67.92 & 9.31 & 78.77 \\
 & ARCQuant & 55.20 & 76.35 & 62.86 & 79.65 & 70.32 & 68.88 & 9.11 & 78.74 \\
 & SharQ & 53.92 & 77.02 & 64.41 & 79.76 & 70.72 & \textbf{69.17} & \textbf{8.96} & \textbf{78.79} \\
\bottomrule
\end{tabular}
\end{table*}
All improvements are obtained without any training, calibration data, or
model-specific tuning, confirming that SharQ operates as a plug-and-play
inference primitive.

\textbf{Multimodal evaluation.}
To assess whether SharQ generalizes beyond text-only language models, we
further evaluate it on Qwen3-VL-8B~\citep{qwen3,qwen2vl}, a vision-language model, across five
standard multimodal benchmarks: VQAv2~\citep{goyal2017makingvvqamatter}, GQA~\citep{gqa}, ScienceQA~\citep{scienceqa}, POPE~\citep{pope}, and TextVQA~\citep{textvqa}. Table~\ref{tab:vlm-results} reports the
results.

\begin{table}[t]
\caption{Multimodal evaluation of SharQ on Qwen3-VL-8B. Accuracy (\%) is
reported for all benchmarks.}
\label{tab:vlm-results}
\centering
\setlength{\tabcolsep}{5pt}
\begin{tabular}{@{}l|ccccc|c@{}}
\toprule
Method & VQA & GQA & SQA & POPE & T-VQA & Avg. \\
\midrule
FP16  & 81.84 & 61.54 & 92.76 & 89.14 & 81.58 & 81.37 \\
\midrule
NVFP4 & 80.47 & 60.58 & 91.58 & 87.73 & 79.47 & 79.97 \\
SharQ & \textbf{81.10} & \textbf{61.04} & \textbf{91.96} & \textbf{88.20} & \textbf{80.57} & \textbf{80.57} \\
\bottomrule
\end{tabular}
\end{table}

SharQ improves over NVFP4 on all five multimodal benchmarks. The average
accuracy rises from 79.97\% to 80.57\%, recovering 43\% of the gap to FP16,
with the largest gain on TextVQA (+1.10\,pp). These results show that the same
online sparse--dense decomposition remains effective for vision-language
architectures.

\subsection{Efficiency}

We evaluate the end-to-end efficiency of SharQ on two representative workloads:
autoregressive language model serving and diffusion-based video generation.

\begin{figure*}[t]
  \centering
  \includegraphics[width=\textwidth]{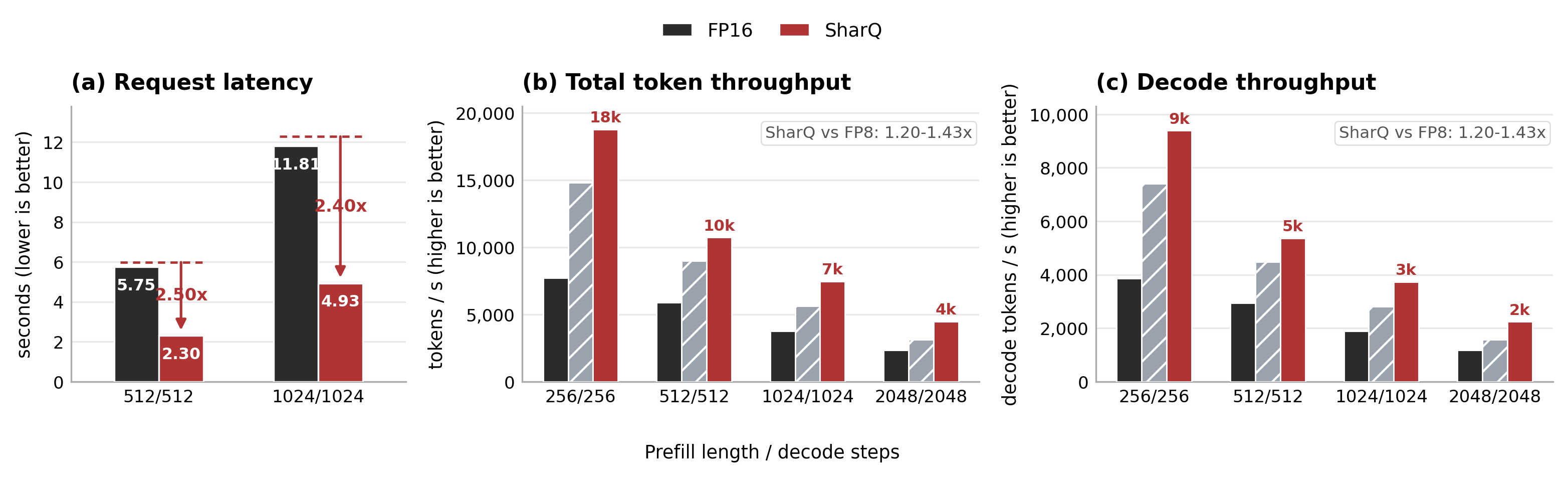}
  \caption{End-to-end serving efficiency of SharQ on RTX 5090 with vLLM (Llama-3.1-8B, batch size 4). Left: request latency; Middle: total token throughput; Right: decode token throughput. SharQ consistently outperforms both FP16 and FP8 baselines.}
  \label{fig:vllm-efficiency}
\end{figure*}

\textbf{Language model serving.}
We integrate SharQ into vLLM~\citep{vllm} and benchmark on an NVIDIA RTX 5090 GPU with
Llama-3.1-8B, comparing against FP16 (full-precision) and FP8
(weight-and-activation quantization) baselines under a fixed batch size of 4,
sweeping input sequence lengths from 256 to 2048 with equal-length decode.
Figure~\ref{fig:vllm-efficiency} summarizes the
results.

SharQ delivers substantial and consistent efficiency gains. Compared to FP16,
it reduces request latency by 2.2--2.4$\times$ and improves total token
throughput by 1.8--2.4$\times$ across the tested sequence lengths. It also
outperforms FP8 by 1.2--1.4$\times$ in total token throughput, reaching
4\,491 total tokens/s versus 3\,146 for FP8 at seqlen\,=\,2048.

\textbf{Video generation.}
To demonstrate the generality of SharQ beyond autoregressive LLMs, we evaluate
it on Wan2.2-T2V-A14B~\citep{wan2025wan}, a state-of-the-art text-to-video diffusion model with
14B parameters. We measure end-to-end generation latency on 4$\times$ RTX 5090
GPUs at 480P and 720P resolutions. We additionally evaluate SharQ combined with
SageAttention~\citep{zhang2025sageattention2efficientattentionthorough} (SharQ+Sage), which applies efficient attention to further
accelerate the diffusion denoising loop. Table~\ref{tab:video-efficiency} reports the results.

\begin{table}[t]
\caption{Video generation latency (seconds, $\downarrow$) on 4$\times$ RTX 5090 (Wan2.2-T2V-A14B).}
\label{tab:video-efficiency}
\centering
\begin{tabular}{@{}l|cc@{}}
\toprule
Method & 480P & 720P \\
\midrule
FP16        & 223.41 & 717.88 \\
SharQ       & 182.21 (\textbf{1.23$\times$}) & 648.04 (\textbf{1.11$\times$}) \\
SharQ+Sage  & \textbf{141.63} (\textbf{1.58$\times$}) & \textbf{457.88} (\textbf{1.57$\times$}) \\
\bottomrule
\end{tabular}
\end{table}

SharQ also generalizes to video generation, reducing latency to 182.21\,s at
480P and 648.04\,s at 720P. Combined with SageAttention, it reaches 1.57--1.58$\times$
speedup over FP16, showing that the method composes naturally with orthogonal
attention-level optimizations.

\subsection{Ablation study}

We conduct two ablation studies to validate the design choices of SharQ.

\textbf{Generalization across FP4 formats.}
Table~\ref{tab:ablation-format} evaluates SharQ on three distinct FP4 numeric
formats: NVFP4~\citep{nvidia2024nvfp4} (with 4:8 in-pairs sparsity), HiF4~\citep{luo2026hifloat4} (with 2:4 sparsity), and
MXFP4~\citep{rouhani2023microscalingdataformatsdeep} (the OCP Microscaling standard). On Qwen3-30B-A3B, SharQ consistently
improves all three formats across every benchmark. For NVFP4, WikiText2
perplexity drops from 9.12 to 8.96 and HellaSwag accuracy rises from 76.76\%
to 77.02\%. For HiF4, the same pattern holds: perplexity improves from 9.08
to 8.92 and HellaSwag from 76.38\% to 77.22\%. The improvement is most
pronounced on MXFP4, the weakest baseline due to its power-of-two scale
constraint and larger group size: SharQ reduces WikiText2 perplexity from
9.70 to 9.19 ($-$0.51) and lifts HellaSwag by +1.36\,pp, Lambada by +2.84\,pp,
WinoGrande by +1.82\,pp, and BoolQ by +0.80\,pp. Across all three formats, BoolQ
accuracy recovers toward the FP16 level, and the gains are consistent in both
direction and magnitude. These results confirm that SharQ's decomposition
strategy is format-agnostic and benefits any block-scaled FP4 format paired
with hardware-supported semi-structured sparsity.

\begin{table}[t]
\caption{Effect of SharQ on different FP4 formats (Qwen3-30B-A3B).
Perplexity ($\downarrow$) for WikiText2.}
\label{tab:ablation-format}
\centering
\begin{tabular}{@{}l|ccccc@{}}
\toprule
Method & Hella & Lamba & Wino & BoolQ & \shortstack{PPL$\downarrow$\\WikiText2}\\
\midrule
FP16        & 77.72 & 64.86 & 70.88  & 88.62 & 8.70 \\
\midrule
NVFP4       & 76.76 & 62.91 & 69.46  & 87.77 & 9.12 \\
+SharQ      & \textbf{77.02} & \textbf{64.41} & \textbf{70.72}  & \textbf{88.41} & \textbf{8.96}  \\
\midrule
HiF4        & 76.38 & 63.28 & 70.17 & 87.61 & 9.08  \\
+SharQ      & \textbf{77.22} & \textbf{64.08} & \textbf{70.24}   & \textbf{88.07} & \textbf{8.92}\\
\midrule
MXFP4       & 74.77 & 59.71 & 67.48 & 86.91 & 9.70 \\
+SharQ      & \textbf{76.13} & \textbf{62.55} & \textbf{69.30}  & \textbf{87.71} & \textbf{9.19}\\
\bottomrule
\end{tabular}
\end{table}

\textbf{Sparse--dense path combinations.}
Table~\ref{tab:ablation-combination} isolates the contribution of each path
in SharQ by varying the combination of sparse and dense computation under NVFP4
on Llama-3.1-8B. Using only the sparse path without residual compensation
(``Sparse'') is catastrophic: WikiText2 perplexity degrades to 82.95 and MMLU
drops to 25.78\%, confirming that discarding all non-outlier activation
information destroys model quality. ``Dense+Dense'' applies two dense NVFP4 paths
and achieves strong accuracy (WikiText2 6.63, MMLU 63.61), but forgoes the
latency benefit of semi-structured sparse computation. ``Sparse+Sparse'' uses two
sparse paths and yields results comparable to the NVFP4 baseline (WikiText2 6.95,
MMLU 61.93), indicating that doubling the sparse computation without a dense
compensation signal does not recover the sparsification loss. SharQ combines a
sparse backbone with a dense residual, achieving WikiText2 6.73 and MMLU 63.76.
This matches or exceeds the accuracy of the fully dense two-path variant while
retaining the sparse execution path for the backbone, validating that the
asymmetric sparse-plus-dense decomposition is the key to SharQ's
accuracy-efficiency trade-off.

\begin{table}[t]
\caption{Ablation on sparse--dense path combinations under NVFP4
(Llama-3.1-8B). ``Sparse'' uses only the sparse path without residual
compensation. ``Dense+Dense'' and ``Sparse+Sparse'' use two paths of the
same type. SharQ combines a sparse backbone with a dense residual.}
\label{tab:ablation-combination}
\centering
\begin{tabular}{@{}l|cccc@{}}
\toprule
Combination & PPL$\downarrow$ & MMLU & BoolQ & ARC-E \\
\midrule
Sparse          & 82.95 & 25.78 & 57.03 & 36.74 \\
Dense + Dense   &  6.63 & 63.61 & 81.22 & 76.98 \\
Sparse + Sparse &  6.95 & 61.93 & 79.45 & 77.61 \\
SharQ           &  6.73 & 63.76 & 80.58 & 77.74 \\
\bottomrule
\end{tabular}
\end{table}

\begin{figure}[t]
  \centering
  \includegraphics[width=\linewidth]{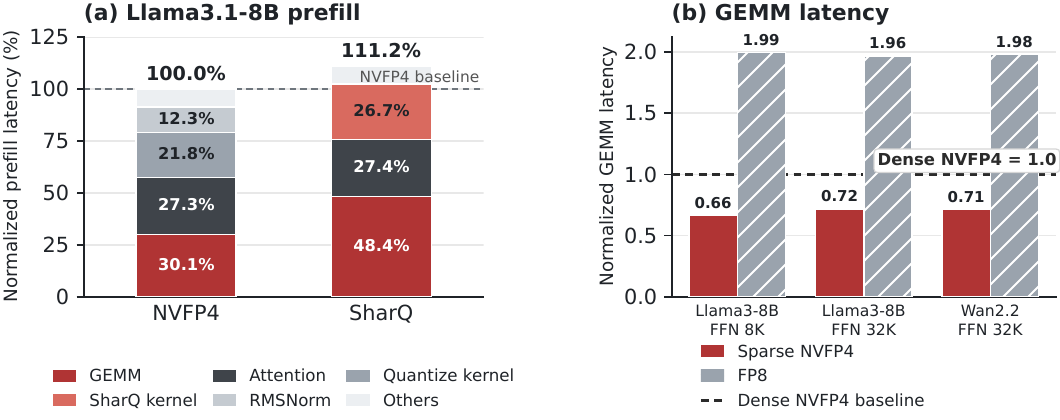}
  \caption{(a)~Normalized prefill latency breakdown on Llama-3.1-8B.
  NVFP4 serves as the 100\% baseline. SharQ replaces the separate
  Quantize and RMSNorm kernels with a single fused SharQ kernel that
  jointly performs quantization, N:M sparse mask generation, and
  normalization.
  (b)~GEMM latency normalized to dense NVFP4 (1.0).
  }
  \label{fig:prefill-breakdown}
\end{figure}

\textbf{Prefill latency breakdown and GEMM speedup.}
Figure~\ref{fig:prefill-breakdown} provides a fine-grained breakdown of the
prefill latency on Llama-3.1-8B and compares per-kernel GEMM latency across
different configurations.

As shown in Figure~\ref{fig:prefill-breakdown}(a), the total prefill latency
of SharQ is 111.2\% relative to NVFP4, representing only an 11.2\% overhead.
This modest increase results from two opposing effects. On one hand, SharQ's
two-path execution (sparse backbone plus dense residual) increases the GEMM
portion from 30.1\% to 48.4\% of the NVFP4 baseline time, as two matrix
multiplications are issued per layer. On the other hand, the fused SharQ
kernel absorbs both the standalone quantization kernel (12.3\% in NVFP4) and
the RMSNorm operator (21.8\% in NVFP4) into a single kernel that jointly
performs FP16-to-FP4 quantization, N:M sparse mask selection and compression,
and layer normalization. This fusion reduces the combined preprocessing cost
from 34.1\% to 26.7\%, saving 7.4 percentage points of prefill time. The net
effect is that SharQ introduces only a small overhead during prefill while
providing substantial accuracy recovery, and its end-to-end latency advantage
manifests primarily in the decode phase where the sparse GEMM path dominates.

Figure~\ref{fig:prefill-breakdown}(b) isolates the GEMM kernel latency,
normalized to dense NVFP4 as 1.0. Sparse NVFP4 achieves 0.66$\times$,
0.72$\times$, and 0.71$\times$ of the dense NVFP4 GEMM latency on
Llama-3.1-8B FFN 8K, Llama-3.1-8B FFN 32K, and Wan2.2 FFN 32K respectively,
corresponding to a 28--34\% GEMM speedup from the 4:8 semi-structured
sparsity. In contrast, FP8 GEMMs require approximately 2$\times$ the latency
of dense NVFP4, confirming that
FP4 formats offer a fundamental compute density advantage over FP8. These
kernel-level measurements validate SharQ's design: the sparse FP4 backbone
exploits hardware-accelerated N:M sparsity for significant arithmetic
savings, while the fused preparation kernel ensures that the overhead of
online decomposition does not erode the throughput gains.

\section{Conclusion}

We presented SharQ, a training-free inference method that combines activation
sparsity with FP4 quantization through an online sparse--dense decomposition.
An input-adaptive N:M mask extracts an outlier-dominated sparse backbone for
hardware-accelerated sparse FP4 computation, and a dense FP4 residual,
defined relative to the quantized sparse backbone, jointly compensates for
mask-induced activation loss and sparse-path quantization error. Three
implementation choices make this decomposition practical: a fused preparation
kernel that absorbs mask generation, residual construction, and layer
normalization into a single operator; shared FP4 weights with path-specific
scale views; and accumulation-based residual compensation that preserves
near-single-kernel data flow. Experiments on dense decoder-only models
(Llama-3.1-8B, Qwen2.5-7B), a mixture-of-experts model (Qwen3-30B-A3B), a
vision-language model (Qwen3-VL-8B), and a video generation model
(Wan2.2-T2V-A14B) show that SharQ consistently recovers a substantial
fraction of FP4 quantization accuracy loss while delivering
2.2--2.4$\times$ latency reduction over FP16 and 1.2--1.4$\times$ throughput
improvement over FP8 in language model serving. The same mechanism
generalizes across NVFP4, HiF4, and MXFP4 without modification, confirming
its format-agnostic nature. SharQ requires no calibration data, retraining,
or model-specific tuning, operating as a plug-and-play inference primitive
wherever block-scaled FP4 and N:M semi-structured sparsity are available.

\bibliographystyle{aaai2026}
\phantomsection\pdfbookmark[0]{References}{references}
\bibliography{refs}

\begin{thebibliography}{49}
\providecommand{\natexlab}[1]{#1}

\bibitem[{Ashkboos et~al.(2024)Ashkboos, Mohtashami, Croci, Li, Cameron, Jaggi, Alistarh, Hoefler, and Hensman}]{ashkboos2024quarot}
Ashkboos, S.; Mohtashami, A.; Croci, M.~L.; Li, B.; Cameron, P.; Jaggi, M.; Alistarh, D.; Hoefler, T.; and Hensman, J. 2024.
\newblock {QuaRot}: Outlier-Free 4-Bit Inference in Rotated {LLMs}.
\newblock In \emph{The Thirty-eighth Annual Conference on Neural Information Processing Systems}.

\bibitem[{Bisk et~al.(2019)Bisk, Zellers, Le~Bras, Gao, and Choi}]{piqa}
Bisk, Y.; Zellers, R.; Le~Bras, R.; Gao, J.; and Choi, Y. 2019.
\newblock {PIQA}: Reasoning about Physical Commonsense in Natural Language.
\newblock arXiv:1911.11641.

\bibitem[{Chen et~al.(2025)Chen, Xi, Zhu, and Chen}]{chen2025oscillationreducedmxfp4trainingvision}
Chen, Y.; Xi, H.; Zhu, J.; and Chen, J. 2025.
\newblock Oscillation-Reduced {MXFP4} Training for Vision Transformers.
\newblock arXiv:2502.20853.

\bibitem[{Clark et~al.(2018)Clark, Cowhey, Etzioni, Khot, Sabharwal, Schoenick, and Tafjord}]{allenai:arc}
Clark, P.; Cowhey, I.; Etzioni, O.; Khot, T.; Sabharwal, A.; Schoenick, C.; and Tafjord, O. 2018.
\newblock Think you have Solved Question Answering? Try {ARC}, the {AI2} Reasoning Challenge.
\newblock arXiv:1803.05457.

\bibitem[{Cook et~al.(2025)Cook, Guo, Xiao, Lin, Wyss, Nazemi, Mishra, del Mundo, Blankevoort, and Han}]{cook2025fouroversix}
Cook, J.; Guo, J.; Xiao, G.; Lin, Y.; Wyss, K.; Nazemi, M.; Mishra, A.; del Mundo, C.; Blankevoort, T.; and Han, S. 2025.
\newblock Four Over Six: More Accurate {NVFP4} Quantization with Adaptive Block Scaling.
\newblock arXiv:2512.02010.

\bibitem[{Darvish~Rouhani et~al.(2023{\natexlab{a}})Darvish~Rouhani, Garegrat, Savell, More, Han, Zhao, Hall, Klar, Chung, Yu, Schulte, Wittig, Bratt, Stephens, Milanovic, Brothers, Dubey, Cornea, Heinecke, Rodriguez, Langhammer, Deng, Naumov, Micikevicius, Siu, and Verrilli}]{mxspecification}
Darvish~Rouhani, B.; Garegrat, N.; Savell, T.; More, A.; Han, K.-N.; Zhao, R.; Hall, M.; Klar, J.; Chung, E.; Yu, Y.; Schulte, M.; Wittig, R.; Bratt, I.; Stephens, N.; Milanovic, J.; Brothers, J.; Dubey, P.; Cornea, M.; Heinecke, A.; Rodriguez, A.; Langhammer, M.; Deng, S.; Naumov, M.; Micikevicius, P.; Siu, M.; and Verrilli, C. 2023{\natexlab{a}}.
\newblock {OCP} Microscaling ({MX}) Specification.
\newblock Open Compute Project.

\bibitem[{Darvish~Rouhani et~al.(2023{\natexlab{b}})Darvish~Rouhani, Zhao, More, Hall, Khodamoradi, Deng, Choudhary, Cornea, Dellinger, Denolf, Dusan, Elango, Golub, Heinecke, James-Roxby, Jani, Kolhe, Langhammer, Li, Melnick, Mesmakhosroshahi, Rodriguez, Schulte, Shafipour, Shao, Siu, Dubey, Micikevicius, Naumov, Verrilli, Wittig, Burger, and Chung}]{rouhani2023microscalingdataformatsdeep}
Darvish~Rouhani, B.; Zhao, R.; More, A.; Hall, M.; Khodamoradi, A.; Deng, S.; Choudhary, D.; Cornea, M.; Dellinger, E.; Denolf, K.; Dusan, S.; Elango, V.; Golub, M.; Heinecke, A.; James-Roxby, P.; Jani, D.; Kolhe, G.; Langhammer, M.; Li, A.; Melnick, L.; Mesmakhosroshahi, M.; Rodriguez, A.; Schulte, M.; Shafipour, R.; Shao, L.; Siu, M.; Dubey, P.; Micikevicius, P.; Naumov, M.; Verrilli, C.; Wittig, R.; Burger, D.; and Chung, E. 2023{\natexlab{b}}.
\newblock Microscaling Data Formats for Deep Learning.
\newblock arXiv:2310.10537.

\bibitem[{Dettmers et~al.(2022)Dettmers, Lewis, Belkada, and Zettlemoyer}]{dettmers2022llmint88bitmatrixmultiplication}
Dettmers, T.; Lewis, M.; Belkada, Y.; and Zettlemoyer, L. 2022.
\newblock {LLM.int8()}: 8-bit Matrix Multiplication for Transformers at Scale.
\newblock arXiv:2208.07339.

\bibitem[{Egiazarian et~al.(2026)Egiazarian, Castro, Kuznedelev, Panferov, Kurtic, Pandit, Marques, Kurtz, Ashkboos, Hoefler, and Alistarh}]{egiazarian2025bridging}
Egiazarian, V.; Castro, R.~L.; Kuznedelev, D.; Panferov, A.; Kurtic, E.; Pandit, S.; Marques, A.; Kurtz, M.; Ashkboos, S.; Hoefler, T.; and Alistarh, D. 2026.
\newblock Bridging the Gap Between Promise and Performance for Microscaling {FP4} Quantization.
\newblock In \emph{The Fourteenth International Conference on Learning Representations (ICLR)}.

\bibitem[{Frantar and Alistarh(2023)}]{sparsegpt}
Frantar, E.; and Alistarh, D. 2023.
\newblock {SparseGPT}: Massive Language Models Can be Accurately Pruned in One-Shot.
\newblock In Krause, A.; Brunskill, E.; Cho, K.; Engelhardt, B.; Sabato, S.; and Scarlett, J., eds., \emph{Proceedings of the 40th International Conference on Machine Learning}, volume 202 of \emph{Proceedings of Machine Learning Research}, 10323--10337. PMLR.

\bibitem[{Frantar et~al.(2023)Frantar, Ashkboos, Hoefler, and Alistarh}]{frantar2023gptqaccurateposttrainingquantization}
Frantar, E.; Ashkboos, S.; Hoefler, T.; and Alistarh, D. 2023.
\newblock {GPTQ}: Accurate Post-Training Quantization for Generative Pre-trained Transformers.
\newblock arXiv:2210.17323.

\bibitem[{Gao et~al.(2024)Gao, Tow, Abbasi, Biderman, Black, DiPofi, Foster, Golding, Hsu, Le~Noac'h, Li, McDonell, Muennighoff, Ociepa, Phang, Reynolds, Schoelkopf, Skowron, Sutawika, Tang, Thite, Wang, Wang, and Zou}]{lm-eval}
Gao, L.; Tow, J.; Abbasi, B.; Biderman, S.; Black, S.; DiPofi, A.; Foster, C.; Golding, L.; Hsu, J.; Le~Noac'h, A.; Li, H.; McDonell, K.; Muennighoff, N.; Ociepa, C.; Phang, J.; Reynolds, L.; Schoelkopf, H.; Skowron, A.; Sutawika, L.; Tang, E.; Thite, A.; Wang, B.; Wang, K.; and Zou, A. 2024.
\newblock The Language Model Evaluation Harness.
\newblock Zenodo, version v0.4.3.

\bibitem[{Goyal et~al.(2017)Goyal, Khot, Summers-Stay, Batra, and Parikh}]{goyal2017makingvvqamatter}
Goyal, Y.; Khot, T.; Summers-Stay, D.; Batra, D.; and Parikh, D. 2017.
\newblock Making the {V} in {VQA} Matter: Elevating the Role of Image Understanding in Visual Question Answering.
\newblock arXiv:1612.00837.

\bibitem[{Grattafiori et~al.(2024)Grattafiori, Dubey, Jauhri, Pandey, Kadian, Al-Dahle, Letman, Mathur et~al.}]{grattafiori2024llama3herdmodels}
Grattafiori, A.; Dubey, A.; Jauhri, A.; Pandey, A.; Kadian, A.; Al-Dahle, A.; Letman, A.; Mathur, A.; et~al. 2024.
\newblock The {Llama} 3 Herd of Models.
\newblock arXiv:2407.21783.

\bibitem[{Hendrycks et~al.(2021)Hendrycks, Burns, Basart, Zou, Mazeika, Song, and Steinhardt}]{mmlu}
Hendrycks, D.; Burns, C.; Basart, S.; Zou, A.; Mazeika, M.; Song, D.; and Steinhardt, J. 2021.
\newblock Measuring Massive Multitask Language Understanding.
\newblock In \emph{International Conference on Learning Representations (ICLR)}.

\bibitem[{Hudson and Manning(2019)}]{gqa}
Hudson, D.~A.; and Manning, C.~D. 2019.
\newblock {GQA}: A New Dataset for Real-World Visual Reasoning and Compositional Question Answering.
\newblock arXiv:1902.09506.

\bibitem[{Kwon et~al.(2023)Kwon, Li, Zhuang, Sheng, Zheng, Yu, Gonzalez, Zhang, and Stoica}]{vllm}
Kwon, W.; Li, Z.; Zhuang, S.; Sheng, Y.; Zheng, L.; Yu, C.~H.; Gonzalez, J.~E.; Zhang, H.; and Stoica, I. 2023.
\newblock Efficient Memory Management for Large Language Model Serving with {PagedAttention}.
\newblock In \emph{Proceedings of the ACM SIGOPS 29th Symposium on Operating Systems Principles}.

\bibitem[{Lee et~al.(2025)Lee, Park, Kim, Kim, Oh, Oh, and Choi}]{lee2025amxfp4tamingactivationoutliers}
Lee, J.; Park, J.; Kim, J.; Kim, Y.; Oh, J.; Oh, J.; and Choi, J. 2025.
\newblock {AMXFP4}: Taming Activation Outliers with Asymmetric Microscaling Floating-Point for 4-bit {LLM} Inference.
\newblock arXiv:2411.09909.

\bibitem[{Li et~al.(2024)Li, Lin, Zhang, Cai, Li, Guo, Xie, Meng, Zhu, and Han}]{li2024svdquant}
Li, M.; Lin, Y.; Zhang, Z.; Cai, T.; Li, X.; Guo, J.; Xie, E.; Meng, C.; Zhu, J.-Y.; and Han, S. 2024.
\newblock {SVDQuant}: Absorbing Outliers by Low-Rank Components for 4-Bit Diffusion Models.
\newblock arXiv:2411.05007.

\bibitem[{Li et~al.(2023)Li, Du, Zhou, Wang, Zhao, and Wen}]{pope}
Li, Y.; Du, Y.; Zhou, K.; Wang, J.; Zhao, W.~X.; and Wen, J.-R. 2023.
\newblock Evaluating Object Hallucination in Large Vision-Language Models.
\newblock arXiv:2305.10355.

\bibitem[{Lin et~al.(2026)Lin, Jia, Xu, Yao, Guo, Wu, Lu, Wei, Zhang, and Sun}]{DuQuant++}
Lin, H.; Jia, X.; Xu, H.; Yao, B.; Guo, X.; Wu, Y.; Lu, Z.; Wei, Y.; Zhang, Q.; and Sun, Z. 2026.
\newblock {DuQuant++}: Fine-grained Rotation Enhances Microscaling {FP4} Quantization.
\newblock arXiv:2604.17789.

\bibitem[{Lin et~al.(2024)Lin, Tang, Tang, Yang, Chen, Wang, Xiao, Dang, Gan, and Han}]{lin2023awq}
Lin, J.; Tang, J.; Tang, H.; Yang, S.; Chen, W.-M.; Wang, W.-C.; Xiao, G.; Dang, X.; Gan, C.; and Han, S. 2024.
\newblock {AWQ}: Activation-aware Weight Quantization for {LLM} Compression and Acceleration.
\newblock In \emph{MLSys}.

\bibitem[{Liu et~al.(2025)Liu, Meng, Luo, Zhang, and Ma}]{liu2025micromix}
Liu, W.; Meng, H.; Luo, Y.; Zhang, P.; and Ma, X. 2025.
\newblock {MicroMix}: Efficient Mixed-Precision Quantization with Microscaling Formats for Large Language Models.
\newblock arXiv:2508.02343.

\bibitem[{Liu et~al.(2023)Liu, Wang, Dao, Zhou, Yuan, Song, Shrivastava, Zhang, Tian, R{\'e} et~al.}]{liu2023deja}
Liu, Z.; Wang, J.; Dao, T.; Zhou, T.; Yuan, B.; Song, Z.; Shrivastava, A.; Zhang, C.; Tian, Y.; R{\'e}, C.; et~al. 2023.
\newblock Deja Vu: Contextual Sparsity for Efficient {LLMs} at Inference Time.
\newblock In \emph{International Conference on Machine Learning}, 22137--22176. PMLR.

\bibitem[{Liu et~al.(2024)Liu, Zhao, Fedorov, Soran, Choudhary, Krishnamoorthi, Chandra, Tian, and Blankevoort}]{liu2024spinquantllmquantizationlearned}
Liu, Z.; Zhao, C.; Fedorov, I.; Soran, B.; Choudhary, D.; Krishnamoorthi, R.; Chandra, V.; Tian, Y.; and Blankevoort, T. 2024.
\newblock {SpinQuant}: {LLM} Quantization with Learned Rotations.
\newblock arXiv:2405.16406.

\bibitem[{Lu et~al.(2022)Lu, Mishra, Xia, Qiu, Chang, Zhu, Tafjord, Clark, and Kalyan}]{scienceqa}
Lu, P.; Mishra, S.; Xia, T.; Qiu, L.; Chang, K.-W.; Zhu, S.-C.; Tafjord, O.; Clark, P.; and Kalyan, A. 2022.
\newblock Learn to Explain: Multimodal Reasoning via Thought Chains for Science Question Answering.
\newblock In \emph{The 36th Conference on Neural Information Processing Systems (NeurIPS)}.

\bibitem[{Luo et~al.(2026)Luo, Huang, Cheng, Yu, Tang, Ma, Wang, Tong, Hu, Xu et~al.}]{luo2026hifloat4}
Luo, Y.; Huang, J.; Cheng, Y.; Yu, Z.; Tang, K.; Ma, X.; Wang, X.; Tong, A.; Hu, G.; Xu, Y.; et~al. 2026.
\newblock {HiFloat4} Format for Language Model Inference.
\newblock arXiv:2602.11287.

\bibitem[{Meng et~al.(2026)Meng, Luo, Zhao, Liu, Zhang, and Ma}]{meng2026arcquant}
Meng, H.; Luo, Y.; Zhao, Y.; Liu, W.; Zhang, P.; and Ma, X. 2026.
\newblock {ARCQuant}: Boosting {NVFP4} Quantization with Augmented Residual Channels for {LLMs}.
\newblock In \emph{Proceedings of the 64th Annual Meeting of the Association for Computational Linguistics (Volume 1: Long Papers)}, 8609--8623. Association for Computational Linguistics.

\bibitem[{Merity et~al.(2016)Merity, Xiong, Bradbury, and Socher}]{wikitext}
Merity, S.; Xiong, C.; Bradbury, J.; and Socher, R. 2016.
\newblock Pointer Sentinel Mixture Models.
\newblock arXiv:1609.07843.

\bibitem[{{NVIDIA}(2024)}]{BlackwellArchitectureTechnical}
{NVIDIA}. 2024.
\newblock {NVIDIA} {Blackwell} Architecture Technical Brief.
\newblock \url{https://resources.nvidia.com/en-us-blackwell-architecture}.

\bibitem[{{NVIDIA Corporation}(2024)}]{nvidia2024nvfp4}
{NVIDIA Corporation}. 2024.
\newblock {cuDNN} Frontend {API} v1.14.0: Block-Scaling Operation.
\newblock \url{https://docs.nvidia.com/deeplearning/cudnn/frontend/v1.14.0/operations/BlockScaling.html}.
\newblock Accessed: 2025-09-16.

\bibitem[{Paperno et~al.(2016)Paperno, Kruszewski, Lazaridou, Pham, Bernardi, Pezzelle, Baroni, Boleda, and Fern{\'a}ndez}]{lambada}
Paperno, D.; Kruszewski, G.; Lazaridou, A.; Pham, N.~Q.; Bernardi, R.; Pezzelle, S.; Baroni, M.; Boleda, G.; and Fern{\'a}ndez, R. 2016.
\newblock The {LAMBADA} Dataset: Word Prediction Requiring a Broad Discourse Context.
\newblock In \emph{Proceedings of the 54th Annual Meeting of the Association for Computational Linguistics (Volume 1: Long Papers)}, 1525--1534. Berlin, Germany: Association for Computational Linguistics.

\bibitem[{Sakaguchi et~al.(2019)Sakaguchi, Le~Bras, Bhagavatula, and Choi}]{sakaguchi2019winograndeadversarialwinogradschema}
Sakaguchi, K.; Le~Bras, R.; Bhagavatula, C.; and Choi, Y. 2019.
\newblock {WinoGrande}: An Adversarial {Winograd} Schema Challenge at Scale.
\newblock arXiv:1907.10641.

\bibitem[{Saxena et~al.(2025)Saxena, Sharify, Roy, and Wang}]{saxena2025resqmixedprecisionquantizationlarge}
Saxena, U.; Sharify, S.; Roy, K.; and Wang, X. 2025.
\newblock {ResQ}: Mixed-Precision Quantization of Large Language Models with Low-Rank Residuals.
\newblock arXiv:2412.14363.

\bibitem[{Shao et~al.(2024)Shao, Chen, Zhang, Xu, Zhao, Li, Zhang, Gao, Qiao, and Luo}]{shao2024omniquantomnidirectionallycalibratedquantization}
Shao, W.; Chen, M.; Zhang, Z.; Xu, P.; Zhao, L.; Li, Z.; Zhang, K.; Gao, P.; Qiao, Y.; and Luo, P. 2024.
\newblock {OmniQuant}: Omnidirectionally Calibrated Quantization for Large Language Models.
\newblock arXiv:2308.13137.

\bibitem[{Shao et~al.(2025)Shao, Wang, Chen, Xu, Wei, and Cheng}]{shao2025block}
Shao, Y.; Wang, P.; Chen, Y.; Xu, C.; Wei, Z.; and Cheng, J. 2025.
\newblock Block Rotation is All You Need for {MXFP4} Quantization.
\newblock arXiv:2511.04214.

\bibitem[{Singh et~al.(2019)Singh, Natarajan, Shah, Jiang, Chen, Batra, Parikh, and Rohrbach}]{textvqa}
Singh, A.; Natarajan, V.; Shah, M.; Jiang, Y.; Chen, X.; Batra, D.; Parikh, D.; and Rohrbach, M. 2019.
\newblock Towards {VQA} Models That Can Read.
\newblock arXiv:1904.08920.

\bibitem[{Song et~al.(2024)Song, Mi, Xie, and Chen}]{song2024powerinfer}
Song, Y.; Mi, Z.; Xie, H.; and Chen, H. 2024.
\newblock {PowerInfer}: Fast Large Language Model Serving with a Consumer-Grade {GPU}.
\newblock In \emph{Proceedings of the ACM SIGOPS 30th Symposium on Operating Systems Principles}, 590--606.

\bibitem[{Sun et~al.(2024{\natexlab{a}})Sun, Chen, Kolter, and Liu}]{sun2024massiveactivationslargelanguage}
Sun, M.; Chen, X.; Kolter, J.~Z.; and Liu, Z. 2024{\natexlab{a}}.
\newblock Massive Activations in Large Language Models.
\newblock arXiv:2402.17762.

\bibitem[{Sun et~al.(2024{\natexlab{b}})Sun, Liu, Bair, and Kolter}]{sun2024simple}
Sun, M.; Liu, Z.; Bair, A.; and Kolter, J.~Z. 2024{\natexlab{b}}.
\newblock A Simple and Effective Pruning Approach for Large Language Models.
\newblock In \emph{The Twelfth International Conference on Learning Representations (ICLR)}.

\bibitem[{Sun et~al.(2025)Sun, Liu, Bai, Bao, Zhao, Li, Hu, Yu, Hou, Yuan, Jiang, Liu, and Yao}]{sun2024flatquant}
Sun, Y.; Liu, R.; Bai, H.; Bao, H.; Zhao, K.; Li, Y.; Hu, J.; Yu, X.; Hou, L.; Yuan, C.; Jiang, X.; Liu, W.; and Yao, J. 2025.
\newblock {FlatQuant}: Flatness Matters for {LLM} Quantization.
\newblock In \emph{Proceedings of the 42nd International Conference on Machine Learning (ICML)}, Proceedings of Machine Learning Research. PMLR.

\bibitem[{Wan et~al.(2025)Wan, Wang, Ai, Wen, Mao, Xie, Chen, Yu, Zhao, Yang et~al.}]{wan2025wan}
Wan, T.; Wang, A.; Ai, B.; Wen, B.; Mao, C.; Xie, C.-W.; Chen, D.; Yu, F.; Zhao, H.; Yang, J.; et~al. 2025.
\newblock Wan: Open and Advanced Large-Scale Video Generative Models.
\newblock arXiv:2503.20314.

\bibitem[{Wang et~al.(2024)Wang, Bai, Tan, Wang, Fan, Bai, Chen, Liu, Wang, Ge, Fan, Dang, Du, Ren, Men, Liu, Zhou, Zhou, and Lin}]{qwen2vl}
Wang, P.; Bai, S.; Tan, S.; Wang, S.; Fan, Z.; Bai, J.; Chen, K.; Liu, X.; Wang, J.; Ge, W.; Fan, Y.; Dang, K.; Du, M.; Ren, X.; Men, R.; Liu, D.; Zhou, C.; Zhou, J.; and Lin, J. 2024.
\newblock {Qwen2-VL}: Enhancing Vision-Language Model's Perception of the World at Any Resolution.
\newblock arXiv:2409.12191.

\bibitem[{Xiao et~al.(2024)Xiao, Lin, Seznec, Wu, Demouth, and Han}]{xiao2024smoothquantaccurateefficientposttraining}
Xiao, G.; Lin, J.; Seznec, M.; Wu, H.; Demouth, J.; and Han, S. 2024.
\newblock {SmoothQuant}: Accurate and Efficient Post-Training Quantization for Large Language Models.
\newblock arXiv:2211.10438.

\bibitem[{Yang et~al.(2025{\natexlab{a}})Yang, Li, Yang, Zhang, Hui, Zheng, Yu, Gao, Huang, Lv, Liu, Zhou, Huang, Ge, Wei, Lin, Tang, Yang, Tu, Zhang, Yang, Yang, Zhou, Lin, Dang, Bao, Yang, Yu, Li, Xue, Zhang, Wang, Zhu, Men, Li, Tang, Ren, Ren, Fan, Su, Zhang, Wan, Liu, Cui, Zhang, and Qiu}]{qwen3}
Yang, A.; Li, A.; Yang, B.; Zhang, B.; Hui, B.; Zheng, B.; Yu, B.; Gao, C.; Huang, C.; Lv, C.; Liu, D.; Zhou, F.; Huang, F.; Ge, H.; Wei, H.; Lin, H.; Tang, J.; Yang, J.; Tu, J.; Zhang, J.; Yang, J.; Yang, J.; Zhou, J.; Lin, J.; Dang, K.; Bao, K.; Yang, K.; Yu, L.; Li, M.; Xue, M.; Zhang, P.; Wang, P.; Zhu, Q.; Men, R.; Li, T.; Tang, T.; Ren, X.; Ren, X.; Fan, Y.; Su, Y.; Zhang, Y.; Wan, Y.; Liu, Y.; Cui, Z.; Zhang, Z.; and Qiu, Z. 2025{\natexlab{a}}.
\newblock {Qwen3} Technical Report.
\newblock arXiv:2505.09388.

\bibitem[{Yang et~al.(2025{\natexlab{b}})Yang, Yang, Zhang, Hui, Zheng, Yu, Li, Liu, Huang, Wei, Lin, Yang, Tu, Zhang, Yang, Yang, Zhou, Lin, Dang, Lu, Bao, Yang, Yu, Li, Xue, Zhang, Zhu, Men, Lin, Li, Tang, Xia, Ren, Ren, Fan, Su, Zhang, Wan, Liu, Cui, Zhang, and Qiu}]{qwen2025qwen25technicalreport}
Yang, A.; Yang, B.; Zhang, B.; Hui, B.; Zheng, B.; Yu, B.; Li, C.; Liu, D.; Huang, F.; Wei, H.; Lin, H.; Yang, J.; Tu, J.; Zhang, J.; Yang, J.; Yang, J.; Zhou, J.; Lin, J.; Dang, K.; Lu, K.; Bao, K.; Yang, K.; Yu, L.; Li, M.; Xue, M.; Zhang, P.; Zhu, Q.; Men, R.; Lin, R.; Li, T.; Tang, T.; Xia, T.; Ren, X.; Ren, X.; Fan, Y.; Su, Y.; Zhang, Y.; Wan, Y.; Liu, Y.; Cui, Z.; Zhang, Z.; and Qiu, Z. 2025{\natexlab{b}}.
\newblock {Qwen2.5} Technical Report.
\newblock arXiv:2412.15115.

\bibitem[{Zellers et~al.(2019)Zellers, Holtzman, Bisk, Farhadi, and Choi}]{zellers2019hellaswag}
Zellers, R.; Holtzman, A.; Bisk, Y.; Farhadi, A.; and Choi, Y. 2019.
\newblock {HellaSwag}: Can a Machine Really Finish Your Sentence?
\newblock arXiv:1905.07830.

\bibitem[{Zhang et~al.(2025)Zhang, Huang, Zhang, Wei, Zhu, and Chen}]{zhang2025sageattention2efficientattentionthorough}
Zhang, J.; Huang, H.; Zhang, P.; Wei, J.; Zhu, J.; and Chen, J. 2025.
\newblock {SageAttention2}: Efficient Attention with Thorough Outlier Smoothing and Per-thread {INT4} Quantization.
\newblock arXiv:2411.10958.

\bibitem[{Zhao et~al.(2024)Zhao, Lin, Zhu, Ye, Chen, Zheng, Ceze, Krishnamurthy, Chen, and Kasikci}]{atom}
Zhao, Y.; Lin, C.-Y.; Zhu, K.; Ye, Z.; Chen, L.; Zheng, S.; Ceze, L.; Krishnamurthy, A.; Chen, T.; and Kasikci, B. 2024.
\newblock Atom: Low-Bit Quantization for Efficient and Accurate {LLM} Serving.
\newblock In Gibbons, P.; Pekhimenko, G.; and De~Sa, C., eds., \emph{Proceedings of Machine Learning and Systems}, volume~6, 196--209.

\end{thebibliography}

\clearpage

\appendix
\section{Theoretical Analysis}

We briefly explain SharQ from two perspectives: distribution and error. Since
SharQ is primarily designed to improve activation representation under a fixed
low-bit weight execution path, we focus on its activation-side compensation
effect.

\textbf{Outlier isolation improves activation quantization friendliness.}
In group-wise FP4 quantization, the local scale is typically determined by the
largest magnitude within each group. A few outliers can therefore reduce the
effective resolution available to ordinary activation values. By extracting
these outlier-dominated components online into the sparse backbone, SharQ
prevents the dense residual path from sharing the same dominant scale with them.
The residual activation thus has a more concentrated local dynamic range and is
better suited to dense low-bit FP4 representation.

\textbf{SharQ converts coupled activation errors into a single residual
reconstruction problem.}
Let the actual activation represented by the sparse path be
\begin{equation}
  \widetilde{\mathbf{X}}_{\mathrm{sp}}
  =
  \mathcal{D}_{\mathrm{sp}}
  \left(
    \mathcal{Q}_{\mathrm{sp}}(\mathbf{X}_{\mathrm{sp}})
  \right).
\end{equation}
SharQ defines the residual as
\begin{equation}
  \mathbf{R}
  =
  \mathbf{X} - \widetilde{\mathbf{X}}_{\mathrm{sp}}.
\end{equation}
Hence, $\mathbf{R}$ contains both the information outside the sparse mask and
the representation error introduced when the retained backbone is
activation-quantized. If the dense path reconstructs the residual as
$\widetilde{\mathbf{R}}$, the final activation reconstruction error satisfies
\begin{equation}
  \mathbf{X} - (\widetilde{\mathbf{X}}_{\mathrm{sp}} + \widetilde{\mathbf{R}})
  =
  \mathbf{R} - \widetilde{\mathbf{R}}.
\end{equation}
This shows that SharQ does not correct sparsification error and quantization
error separately; it converts them into a single residual reconstruction
problem. Furthermore, under a fixed weight operator $\bar{\mathbf{W}}$, the
output perturbation induced by activation approximation satisfies
\begin{equation}
  \| \mathbf{X}\bar{\mathbf{W}} - (\widetilde{\mathbf{X}}_{\mathrm{sp}} + \widetilde{\mathbf{R}})\bar{\mathbf{W}} \|_F
  \le
  \| \mathbf{R} - \widetilde{\mathbf{R}} \|_F \, \| \bar{\mathbf{W}} \|_2.
\end{equation}
Therefore, the key role of SharQ is not to make the sparse path reconstruct all
activations accurately by itself, but to let the residual path jointly absorb
sparse approximation loss and activation quantization loss.

\section{Extended Related Work Discussion}

\subsection{Quantization}

Post-Training Quantization (PTQ) is pivotal for efficient LLM inference. Early work identified activation outliers as the primary low-bit bottleneck: LLM.int8()~\citep{dettmers2022llmint88bitmatrixmultiplication} introduced mixed-precision decomposition to isolate outlier channels at INT8 precision. Since then, methods have advanced from multiple complementary angles.

\textbf{Transformation-based methods} reshape activation distributions to suppress outliers before quantization. SmoothQuant~\citep{xiao2024smoothquantaccurateefficientposttraining} migrates quantization difficulty from activations to weights via per-channel scaling, proving effective for INT8. For 4-bit scenarios, QuaRot~\citep{ashkboos2024quarot} and SpinQuant~\citep{liu2024spinquantllmquantizationlearned} apply orthogonal rotations to flatten value distributions, while FlatQuant~\citep{sun2024flatquant} further learns affine or orthogonal-scaling transforms to better fit quantization-friendly distributions. However, recent analysis shows that global transformations may disrupt local block statistics critical for fine-grained formats like NVFP4~\citep{shao2025block}, limiting their effectiveness in block-scaled FP4 regimes.

\textbf{Mixed-precision methods} retain sensitive information in higher precision. Atom~\citep{atom} adopts a selection-based strategy, keeping a subset of outlier channels in INT8 or FP16 while quantizing the bulk to low-bit. Beyond simple channel selection, decomposition-based approaches such as ResQ~\citep{saxena2025resqmixedprecisionquantizationlarge} and SVDQuant~\citep{li2024svdquant} extract outliers into a separate high-precision low-rank branch, leaving the residual bulk in 4-bit. MicroMix~\citep{liu2025micromix} explores mixing MXFP4 with MXFP8 on the Blackwell architecture. However, mixed-precision approaches often conflict with hardware constraints on unified-precision Tensor Core computation.

\textbf{Compensation-based methods} minimize reconstruction error via optimization. GPTQ~\citep{frantar2023gptqaccurateposttrainingquantization} utilizes Hessian-based optimization to adjust quantized weights. OmniQuant~\citep{shao2024omniquantomnidirectionallycalibratedquantization} omnidirectionally calibrates quantization parameters. AWQ~\citep{lin2023awq} leverages activation statistics to protect salient weight channels. ARCQuant~\citep{meng2026arcquant} targets NVFP4 specifically by augmenting the activation matrix with quantized residual channels, enabling error compensation within a single unified GEMM call.

Despite these advances, mainstream quantization pipelines continue to treat activations as dense tensors to be compressed. Few works jointly exploit the observed activation sparsity as a dedicated hardware-accelerated computation path alongside quantization, leaving significant potential for further inference acceleration.

\subsection{Sparsity}

\textbf{Weight sparsity} has been extensively studied for LLM compression. SparseGPT~\citep{sparsegpt} achieves one-shot unstructured and semi-structured (2:4, 4:8) pruning on massive LLMs via approximate Hessian-based reconstruction. Wanda~\citep{sun2024simple} simplifies this with a pruning criterion combining weight magnitude and input activation norms. These methods learn or calibrate a fixed mask offline and reuse it for all inputs, making them well-suited for static weight compression but inapplicable to dynamic activations.

\textbf{Activation sparsity} exploits the observation that large-magnitude activations are sparse and carry disproportionate output contribution. DejaVu~\citep{liu2023deja} employs context-dependent prediction to identify and skip low-impact neurons at inference time. PowerInfer~\citep{song2024powerinfer} exploits neuron activation locality for efficient GPU/CPU hybrid serving. However, these methods typically rely on coarse-grained or unstructured sparsity patterns, and often require continued training, prediction modules, or architectural modifications to achieve meaningful acceleration.

\textbf{Online semi-structured activation sparsity}, the paradigm SharQ adopts, remains relatively unexplored. While N:M semi-structured sparsity has been widely adopted for weight compression under hardware constraints, with works such as SparseGPT~\citep{sparsegpt} systematically exploring mask optimization under 2:4 and 4:8 structural constraints, its \emph{plug-and-play} adaptation on the activation side is limited. The key challenge is that activation masks must be generated online from the current input at minimal cost, while satisfying strict hardware patterns (e.g., 4:8 in-pairs for NVFP4). Direct application of dynamic N:M masks typically causes substantial accuracy degradation, as moderate activation values outside the mask are discarded. SharQ addresses this through online sparse--dense decomposition, which treats sparsification as routing rather than dropping, and jointly compensates for both mask-induced loss and sparse-path quantization error via a dense residual path.

\subsection{Fine-grained numeric formats and hardware support}
As compute-efficiency demands intensify, fine-grained low-bit numeric formats have become the standard paradigm for LLM deployment. The Open Compute Project (OCP) Microscaling (MX) specification~\citep{rouhani2023microscalingdataformatsdeep} defines a block-shared scaling representation, where MXFP4 balances precision and storage at ultra-fine granularity. On the hardware side, the NVIDIA Blackwell architecture~\citep{BlackwellArchitectureTechnical} natively supports FP4 inference covering both NVFP4 and MXFP4 in its fifth-generation Tensor Cores, delivering multi-fold throughput gains over previous generations. HiF4~\citep{luo2026hifloat4} proposes a three-level hierarchical scaling design that achieves wider dynamic range (69 binades) with lower hardware area cost.

To align with these hardware advances, several algorithmic adaptations~\citep{chen2025oscillationreducedmxfp4trainingvision,DuQuant++,lee2025amxfp4tamingactivationoutliers} have emerged targeting FP4-specific challenges. ARCQuant~\citep{meng2026arcquant} targets NVFP4 with augmented residual channels that compensate representation loss while strictly adhering to hardware-friendly compute organization. MR-GPTQ~\citep{egiazarian2025bridging} introduces a Micro-Rotated variant of GPTQ tailored to FP4's unique properties, using block-wise Hadamard transforms and format-specific optimizations to significantly boost both MXFP4 and NVFP4 accuracy with high-performance GPU kernels achieving up to 2.2$\times$ end-to-end speedup on B200. Four Over Six~\citep{cook2025fouroversix} proposes adaptive block scaling for NVFP4 that reduces quantization error on near-maximal values by making the distribution of representable FP4 values more uniform. SVDQuant~\citep{li2024svdquant} absorbs activation outliers via low-rank components for 4-bit diffusion models. MicroMix~\citep{liu2025micromix} jointly models Microscaling with mixed precision, dynamically assigning format granularity across tensor blocks.

These works collectively demonstrate that NVFP4 and related Microscaling formats constitute a central research direction for next-generation LLM inference. However, existing methods still predominantly encode activations as dense tensors, leaving the synergistic acceleration potential of activation sparsity and hardware-native N:M semi-structured patterns within fine-grained FP4 compute pipelines largely unexplored, precisely the gap that SharQ addresses.

\section{NVFP4 Data Format}
\label{app:nvfp4}

This section describes the NVFP4 block floating-point format used as the
primary quantization target in our experiments. NVFP4 is a proprietary format
introduced with the NVIDIA Blackwell architecture that improves upon the OCP MXFP4 standard by addressing
two key limitations of prior block-scaled FP4 designs.

\subsection{Design rationale}

MXFP4 (OCP Microscaling FP4) uses E2M1 elements with a shared power-of-two
8-bit exponent (E8M0) and a group size of 32, yielding an average storage cost
of 4.25 bits per value. However, the power-of-two constraint on the block scale
cannot guarantee normalization of each group's peak magnitude to the
representable upper bound of E2M1, wasting intra-group dynamic range. NVFP4
introduces two changes to address this:

\begin{enumerate}
\item \textbf{Floating-point block scale.} NVFP4 replaces the power-of-two
E8M0 scale with a fine-grained FP8-E4M3 floating-point scale. This allows the
scale to normalize group peaks more precisely to the upper bound of E2M1,
fully utilizing the intra-group expressive space.

\item \textbf{Smaller group size.} NVFP4 adopts a 16-element group size
(reduced from 32 in MXFP4), raising the average storage cost from 4.25 to 4.5
bits per value but effectively reducing outlier-driven quantization error by
providing finer-grained scaling granularity.
\end{enumerate}

\subsection{Format structure}

An NVFP4 block consists of 16 four-bit E2M1 elements sharing a single FP8-E4M3
block scale, for a total storage of $16 \times 4 + 8 = 72$ bits, or 4.5 bits
per value. The E2M1 element format allocates 2 bits to the exponent and 1 bit
to the mantissa (with an implicit leading 1 for normal values), plus 1 sign
bit. The representable nonzero magnitudes in E2M1 are
$\{0.5, 1.0, 1.5, 2.0, 3.0, 4.0, 6.0\}$, with the maximum value being 6 and
special encodings reserved for $\pm 0$ and NaN.

The FP8-E4M3 block scale uses 4 exponent bits (bias 7) and 3 mantissa bits,
providing a scale dynamic range of $[-10, 11]$ in binades (22 binades total).
Combined with the E2M1 element range, the local intra-group dynamic range is
$\log_2(6/0.5) = 3.58$ binades.

\subsection{Per-tensor scaling}

Because the E4M3 block scale provides only 22 binades of global dynamic range,
NVFP4 requires an additional software-based per-tensor scaling (PTS) step
before format conversion. PTS normalizes each tensor's peak magnitude to a
target value (typically 2688 for activations) so that the E4M3 scale range is
sufficient to cover the tensor's value distribution. This additional
preprocessing step incurs some overhead during inference but is necessary to
prevent overflow and underflow in the block scale.

\subsection{Sparsity support}

On the Blackwell architecture, NVFP4 natively supports 4:8 semi-structured
sparsity in pairs. Within each group of 8 consecutive elements (organized as 4
pairs), the hardware retains 4 elements (2 pairs) and prunes the remaining 4.
This ``in-pairs'' constraint means that sparsity selection operates on pairs of
adjacent elements rather than individual values, which simplifies hardware
metadata encoding while still enabling effective activation sparsity. SharQ
exploits this 4:8 in-pairs pattern as the sparse backbone constraint in its
NVFP4 instantiation.

\subsection{Comparison with other FP4 formats}

Table~\ref{tab:nvfp4-comparison} summarizes the key differences between NVFP4
and other block-scaled FP4 formats.

\begin{table}[t]
\caption{Comparison of block-scaled FP4 format designs.}
\label{tab:nvfp4-comparison}
\centering
\small
\begin{tabular}{@{}lccc@{}}
\toprule
Feature & MX4 & MXFP4 & NVFP4 \\
\midrule
4-bit element        & S1P1       & E2M1  & E2M1 \\
Scale format         & E8M0       & E8M0  & E4M3 \\
Group size           & 16         & 32    & 16 \\
Storage (bits)       & 4.0        & 4.25  & 4.5 \\
Scale type           & Power-of-2 & Power-of-2 & Float \\
Precision (bits)     & 2          & 2     & 2 \\
Local range          & 2.81 bin.  & 3.58 bin. & 3.58 bin. \\
Global range         & N/A        & N/A   & 22 bin. \\
Requires PTS         & No         & No    & Yes \\
\bottomrule
\end{tabular}
\end{table}

NVFP4 achieves better quantization accuracy than MXFP4 by combining the
floating-point scale (which fully utilizes E2M1's representable range) with a
smaller group size (which reduces the impact of outliers within each block).
However, the limited global dynamic range of E4M3 necessitates per-tensor
scaling, and the smaller group size increases metadata overhead. These
trade-offs position NVFP4 as a format optimized for inference accuracy on
current hardware, while formats like HiF4 (Appendix~\ref{app:hif4}) explore
alternative designs with wider dynamic range and lower hardware cost.

\section{HiFloat4 (HiF4) Data Format}
\label{app:hif4}

This section provides a detailed description of the HiFloat4 (HiF4) block
floating-point format, which is one of the FP4 numeric formats evaluated in our
ablation study (Table~\ref{tab:ablation-format}). HiF4 was proposed as a
high-fidelity 4-bit BFP format for deep learning inference and training. We
summarize its structure, encoding, and key differences from NVFP4 to make the
paper self-contained.

\subsection{Format overview}

A basic HiF4 unit packs 64 four-bit in-group elements together with 32 bits of
shared scaling metadata, resulting in an average storage cost of 4.5 bits per
value (identical to NVFP4). The distinguishing feature of HiF4 is a three-level
scaling hierarchy that captures both inter-group and intra-group dynamic range
variation, improving representational utilization compared with single- or
two-level scaling designs.

\subsection{Three-level scaling metadata}

The 32-bit scaling metadata is organized into three hierarchical levels:

\textbf{Level-1: Global base scale (E6M2).}
The first level is a specially designed unsigned 8-bit floating-point format
called E6M2. It assigns 6 bits to the exponent field with a bias of 48 and 2
bits to the mantissa field with one hidden integer bit set to~1. E6M2 encodes
NaN (Not a Number) but does not support infinity or zero; only normal mode is
used. If we denote the unbiased exponent as $E$ and the mantissa as $M$, an
E6M2 value is interpreted as
\begin{equation}
  X = 2^{E} \times 1.M .
  \label{eq:e6m2}
\end{equation}
This scale provides a wide dynamic range across groups (unbiased exponent range
$[-48, 15]$) and normalizes each group's peak magnitude to the representable
upper bound of the remaining hierarchical structure, ensuring full utilization
of the intra-group dynamic range.

\textbf{Level-2: 8-way one-bit micro-exponents (E1\_8).}
The second level consists of 8 one-bit micro-exponents, each shared by 8
consecutive in-group elements. Each E1 bit encodes either 1 or 0, representing
a very fine-grained power-of-two scaling factor ($2^{E1}$ collapses to either
$2$ or $1$). Each level-2 E1 further connects to two adjacent level-3
micro-exponents.

\textbf{Level-3: 16-way one-bit micro-exponents (E1\_16).}
The third level consists of 16 one-bit micro-exponents, each shared by 4
contiguous four-bit elements within its local group. Together with the level-2
micro-exponents, these refine the intra-group dynamic range by capturing local
exponent differences, effectively mitigating the impact of outliers and
suppressing quantization error.

All three levels of scaling metadata occupy 32 bits in total ($8 + 8 + 16$),
distributed over 64 in-group elements, leading to an extra overhead of 0.5 bits
per value.

\subsection{Element encoding: S1P2}

HiF4 encodes the 64 four-bit in-group elements using the sign-magnitude S1P2
representation. In the SXPY notation, S denotes the sign bit and P indicates
the binary point; the value X preceding P designates the integer part, and the
value Y following P designates the fractional part. Conceptually, S1P2 is
equivalent to the E1M2 format in floating-point representation. The encoding
details are summarized in Table~\ref{tab:hif4-encoding}.

\begin{table}[t]
\caption{E6M2 and S1P2 encoding details used in HiF4.}
\label{tab:hif4-encoding}
\centering
\small
\begin{tabular}{@{}lcc@{}}
\toprule
Property & FP8-E6M2 & S1P2 \\
\midrule
Exp bias      & 48            & N/A \\
Unbiased exp  & $[-48,\;15]$  & N/A \\
Infinity      & N/A           & N/A \\
Zero          & N/A           & $\text{S}0.00_2 = \pm 0.00$ \\
NaN           & $111111\_11_2$ & N/A \\
Max value     & $111111\_10_2 = 2^{15} \times 1.50$ & $\text{S}1.11_2 = \pm 1.75$ \\
Min value     & $000000\_00_2 = 2^{-48} \times 1.00$ & $\text{S}0.01_2 = \pm 0.25$ \\
\bottomrule
\end{tabular}
\end{table}

\subsection{Value representation}

Based on the three-level hierarchy, each of the 64 real numbers
$\{V_i\}_{i=1}^{64}$ in a HiF4 group can be expressed as follows. If
$\text{E6M2} = \text{NaN}$, then $V_i = \text{NaN}$ for all
$i \in [1, 64]$. Otherwise,
\begin{equation}
  V_i
  =
  \text{E6M2}
  \times
  2^{\left(\{\text{E1\_8}\}_{\lceil i/8 \rceil}
    + \{\text{E1\_16}\}_{\lceil i/4 \rceil}\right)}
  \times
  \{\text{S1P2}\}_i ,
  \label{eq:hif4-value}
\end{equation}
where $\{\text{E1\_8}\}_j$ ($j \in [1,8]$) and
$\{\text{E1\_16}\}_k$ ($k \in [1,16]$) denote the level-2 and level-3
micro-exponents, respectively. The available intra-group dynamic range of HiF4
is $\log_2(7/0.25) = 4.81$ binades, since the maximum positive value within a
group is $2^{(1+1)} \times 1.75 = 7$ and the minimum positive value is
$2^{(0+0)} \times 0.25 = 0.25$.

\subsection{Comparison with NVFP4}

Table~\ref{tab:hif4-vs-nvfp4} compares the key features of HiF4 and NVFP4.
Despite identical average storage costs, HiF4 offers a substantially wider
global dynamic range (69 vs.\ 22 binades), higher significand precision (3 bits
vs.\ 2 bits), and a larger local dynamic range (4.81 vs.\ 3.58 binades). These
properties make HiF4 more robust to outlier-driven quantization error and
reduce the need for software-based per-tensor scaling (PTS) that NVFP4
typically requires.

\begin{table}[t]
\caption{Feature comparison between HiF4 and NVFP4.}
\label{tab:hif4-vs-nvfp4}
\centering
\small
\begin{tabular}{@{}lcc@{}}
\toprule
Feature & HiF4 & NVFP4 \\
\midrule
Storage          & 4.5 bits/value & 4.5 bits/value \\
Group size       & 64             & 16 \\
Special values   & NaN and $\pm 0$ & NaN and $\pm 0$ \\
4-bit element    & S1P2 (E1M2)   & E2M1 \\
Precision        & 3 bits         & 2 bits \\
Base scale       & E6M2           & E4M3 \\
Max value        & $2^{18} \times 1.3125$ & $2^{11} \times 1.3125$ \\
Min value        & $2^{-50}$      & $2^{-10}$ \\
Global range     & $[-50,\;18]$: 69 bin. & $[-10,\;11]$: 22 bin. \\
Local range      & 4.81 bin.      & 3.58 bin. \\
\bottomrule
\end{tabular}
\end{table}

\subsection{Conversion from BF16 to HiF4}

Algorithm~\ref{alg:bf16-to-hif4} outlines the conversion procedure from a
64-length BF16 vector to a HiF4 unit. The algorithm proceeds in three stages:
(1)~a three-level tree reduction to find local and global peak magnitudes;
(2)~derivation of the three-level hierarchical scaling metadata (E6M2, E1\_8,
E1\_16); and (3)~scaling and quantization of the 64 in-group elements into S1P2
format. All rounding operations use round-half-to-even or
round-half-away-from-zero. The E6M2 reciprocal computation can be efficiently
implemented using only a 4-entry lookup table indexed by the 2-bit mantissa,
combined with simple exponent subtraction.

\begin{algorithm}[!htb]
\caption{Conversion from BF16 to HiF4}
\label{alg:bf16-to-hif4}
\small
\begin{algorithmic}[1]
\REQUIRE A 64-length BF16 vector $V64$
\ENSURE A HiF4 unit: E6M2, E1\_8, E1\_16, S1P2\_64
\STATE \textbf{// Stage 1: Three-level tree reduction}
\FOR{$i = 1$ to $16$}
  \STATE $V16[i] = \max(|V64[4 \times i - 3 : 4 \times i]|)$
\ENDFOR
\FOR{$i = 1$ to $8$}
  \STATE $V8[i] = \max(V16[2 \times i - 1 : 2 \times i])$
\ENDFOR
\STATE $Vmax = \max(V8)$
\STATE \textbf{// Stage 2: Derive scaling metadata}
\STATE $SF_{BF16} = Vmax \times (1/7)_{BF16}$
  \hfill\COMMENT{High-precision scale factor}
\STATE $\text{E6M2} = \text{BF16\_to\_E6M2}(SF_{BF16})$
  \hfill\COMMENT{Level-1 scale}
\STATE $\text{E6M2\_REC} = \text{E6M2\_REC\_to\_BF16}(\text{E6M2})$
  \hfill\COMMENT{Reciprocal}
\STATE $\text{E1\_8}[i] = (V8[i] \times \text{E6M2\_REC} \geq 4)\ ?\ 1 : 0$
  \hfill\COMMENT{Level-2 scales}
\FOR{$i = 1$ to $16$}
  \STATE $\text{E1\_16}[i] = (V16[i] \times \text{E6M2\_REC} \times 2^{-\text{E1\_8}[\lceil i/2 \rceil]} \geq 2)$\\
    \hspace*{1.5em}$?\ 1 : 0$
\ENDFOR
\STATE \textbf{// Stage 3: Quantize 64 elements}
\FOR{$i = 1$ to $64$}
  \STATE $V64\_\text{scaled}[i] = V64[i] \times \text{E6M2\_REC}$\\
    \hspace*{1.5em}$\times\, 2^{-\text{E1\_8}[\lceil i/8 \rceil]} \times 2^{-\text{E1\_16}[\lceil i/4 \rceil]}$
\ENDFOR
\STATE $\text{S1P2\_64} = \text{BF16\_to\_S1P2}(V64\_\text{scaled})$
  \hfill\COMMENT{Quantize to S1P2}
\end{algorithmic}
\end{algorithm}

For the HiF4 dot product, level-2 and level-3 micro-exponents can be
implemented as simple left-shift operations in hardware. When integrated into
existing 64-length dot-product units originally optimized for 16-bit and 8-bit
formats, HiF4 occupies only approximately one-third the incremental area of
NVFP4 and reduces power consumption by about 10\%, enabling a more area- and
power-efficient implementation for matrix multiplication.

\end{document}